\documentclass[conference]{IEEEtran}
\IEEEoverridecommandlockouts

\usepackage[comma,sort&compress,numbers]{natbib}
\usepackage{amsmath,amssymb,amsfonts}
\usepackage{algorithmic}
\usepackage{soul}
\usepackage{graphicx}
\usepackage{textcomp}
\usepackage{xcolor}
\usepackage{multirow}
\usepackage{booktabs}
\usepackage{arydshln}
\usepackage{hyperref}
\usepackage{subcaption}
\usepackage[font=small]{caption}
\usepackage{flushend}

\def\BibTeX{{\rm B\kern-.05em{\sc i\kern-.025em b}\kern-.08em
    T\kern-.1667em\lower.7ex\hbox{E}\kern-.125emX}}

    \title{Ultra-lightweight Neural Video Representation Compression
    \thanks{This work was funded by UK EPSRC (iCASE Awards), BT and the UKRI MyWorld Strength in Places Programme (SIPF00006/1). High performance computational facilities were provided by the Advanced Computing Research Centre and Bristol Digital Futures Institute, University of Bristol.}}

\author{Ho Man Kwan\textsuperscript{\dag}, Tianhao Peng\textsuperscript{\dag}, Ge Gao\textsuperscript{\dag}, Fan Zhang\textsuperscript{\dag}, Mike Nilsson\textsuperscript{\S}, Andrew Gower\textsuperscript{\S}, David Bull\textsuperscript{\dag}\\
\textsuperscript{\dag}\textit{Visual Information Laboratory, University of Bristol, UK} \\
\texttt{\{hm.kwan, tianhao.peng, ge1.gao, fan.zhang, dave.bull\}@bristol.ac.uk}\\
\textsuperscript{\S}\textit{Network Connectivity Services Research, BT, UK}\\
\texttt{\{mike.nilsson, andrew.p.gower\}@bt.com}}

\begin{document}
\maketitle

\begin{abstract}
Recent works have demonstrated the viability of utilizing over-fitted implicit neural representations (INRs) as alternatives to autoencoder-based models for neural video compression. Among these INR-based video codecs, Neural Video Representation Compression (NVRC) was the first to adopt a fully end-to-end compression framework that compresses INRs, achieving state-of-the-art performance. Moreover, some recently proposed lightweight INRs have shown comparable performance to their baseline codecs with computational complexity lower than 10kMACs/pixel. In this work, we extend NVRC toward lightweight representations, and propose NVRC-Lite, which incorporates two key changes. Firstly, we integrated multi-scale feature grids into our lightweight neural representation, and the use of higher resolution grids significantly improves the performance of INRs at low complexity. Secondly, we address the issue that existing INRs typically leverage autoregressive models for entropy coding: these are effective but impractical due to their slow coding speed. In this work, we propose an octree-based context model for entropy coding high-dimensional feature grids, which accelerates the entropy coding module of the model. Our experimental results demonstrate that NVRC-Lite outperforms C3, one of the best lightweight INR-based video codecs, with up to 21.03\% and 23.06\% BD-rate savings when measured in PSNR and MS-SSIM, respectively, while achieving 8.4$\times$ encoding and 2.5$\times$ decoding speedup. The implementation of NVRC-Lite will be made available at \href{https://github.com}{here}.
\end{abstract}


\section{Introduction}\label{sec:introduction}

Driven by the rapid increase in internet video traffic and improvements in video quality, the demand for more efficient video codecs continues to grow \cite{bull2021intelligent}. While conventional video coding algorithms, based on classical signal processing and information theories \cite{hm,vtm,han2021technical}, have benefited from decades of research, neural video codecs that leverage deep learning techniques have only recently emerged as a promising alternative for building an end-to-end framework. Some recent contributions of this type have been reported to outperform conventional codecs \cite{xiang2022mimt,chen2023b,li2023neural,qi2024long}.

Following early advances in neural image compression \cite{balle2017endtoend,ballé2018variational},  neural video codecs \cite{lu2019dvc,rippel2019learned,habibian2019video} often employ autoencoders to implement transform coding, replacing traditional linear transformations by non-linear processes learned by neural networks. Subsequent learned video codecs further exploited temporal redundancy, using techniques such as inter-frame coding with diverse coding structures \cite{li2021deep,li2023neural,gao2025pnvc,sheng2025bi}, 3D autoencoders which extend their 2D counterparts \cite{habibian2019video,gao2025givic}, and error-propagation-aware regimes \cite{lu2020content,qi2023motion}.

In most cases, optimizing a neural video codec requires extensive offline training on a large-scale dataset to ensure model generalization. This requires the codec to possess high model capacity, in order to enable sufficient generalization for encoding a wide range of video content within a single model. However, in practice, there remains a discrepancy between the distribution of the training set and the testing video instance \cite{yang2020improving}, which may result in suboptimal performance. This can be addressed in two ways: fine-tuning a pre-trained, generic model on the input video \cite{vanrozendaal2023instanceadaptive,chen2024group,gao2025pnvc}, or learning a compact representation entirely from scratch, which is usually achieved using Implicit Neural Representations (INRs) \cite{sitzmann2020implicit}.

When INRs are used for video coding, the neural network is trained to map coordinates to frames \cite{chen2021nerv} or patches \cite{bai2023ps,kwan2023hinerv}, yielding bespoke quantized weights as a compressed representation of the video instance. As this type of instance-specific model only needs to be overfitted for one video, it is substantially more compact than other generalized neural video codecs, resulting in a dramatic reduction in decoder complexity. The coding performance of recent INR-based video compression methods has been significantly improved \cite{li2022nerv,maiya2023nirvana,chen2023hnerv,kwan2023hinerv,kwan2024immersive}, with the state of the art, NVRC \cite{kwan2024nvrc}, outperforming both the latest standard video codecs (e.g., VVC VTM \cite{vtm} in Random Access mode) and autoencoder-based neural video codecs \cite{li2023neural}. While these models are efficient in decoding, some recent studies \cite{leguay2024cool, kim2024c3} have demonstrated the potential to maintain reasonable coding efficiency at a significantly lower computational budget (specifically, under 10kMACs/pixel). This highlights a new research direction for ultra-lightweight neural video codecs.

Inspired by these advancements, this paper proposes a principled redesign of the NVRC architecture, tailored specifically to the ultra-lightweight regime. Firstly, we employ a lightweight model with multi-scale feature grids as our neural representation, demonstrating that the use of higher-resolution grids is essential in improving reconstruction quality at very low complexity. Secondly, we address the relatively slow entropy coding speed of existing INR-based codecs, which typically employ a slow autoregressive model. We replace this with a novel octree-based context model specifically designed for entropy coding high-dimensional feature grids, which further accelerates the (decoding) inference speed. 

The proposed NVRC-Lite codec has been tested on two commonly used datasets, UVG \cite{mercat2020uvg} and HEVC-B \cite{bossen2010common}, against three benchmark methods. The results demonstrate its excellent tradeoff between computational complexity (in both encoding and decoding) and coding performance. Specifically, it achieves average BD-rate savings of 21.03\% and 23.06\% (in PSNR and MS-SSIM, respectively) against one of the best lightweight INR-based video codecs, C3 \cite{kim2024c3}, with 8.4$\times$ encoding and 2.5$\times$ decoding speedup. 

\section{Method}\label{sec:method}

\begin{figure}[!t]
    \centering
    \includegraphics[width=1.0\linewidth]{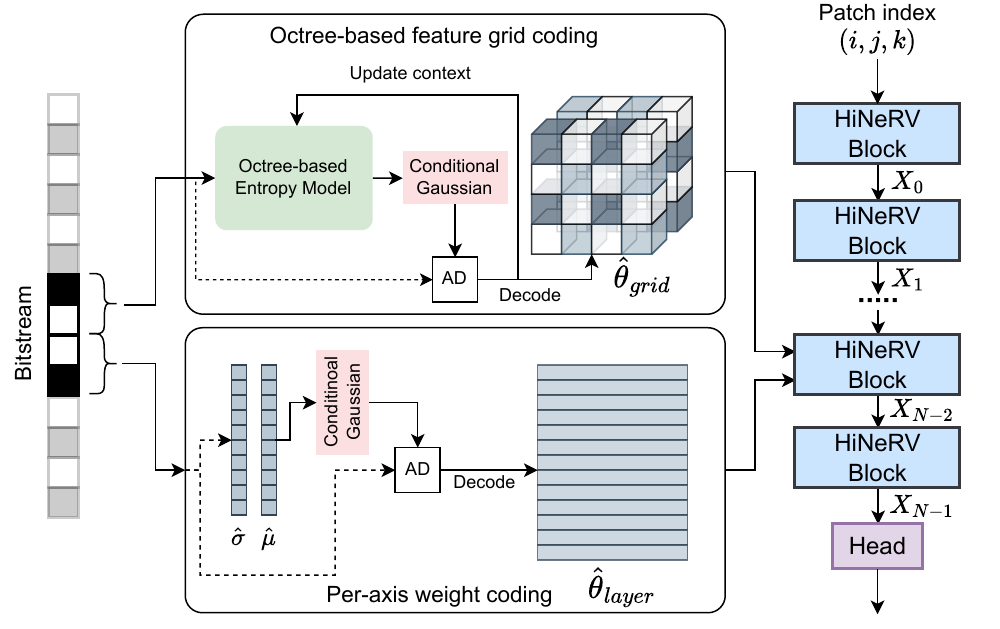}
    \caption{The proposed NVRC-Lite framework. It contains an INR (HiNeRV\cite{kwan2023hinerv} in this case) with feature inputs to multiple blocks at different resolutions, and performs parameter coding by utilizing the proposed octree model for efficient entropy coding. Note that NVRC-Lite also follows NVRC \cite{kwan2024nvrc} to perform parameter coding in a hierarchical manner (details omitted here).}
    \label{fig:framework}
    \vspace{-5pt}
\end{figure}

The proposed NVRC-Lite, as illustrated in \autoref{fig:framework}, is a lightweight video representation coding framework based on NVRC \cite{kwan2024nvrc}, the state-of-the-art INR-based video codec incorporating end-to-end optimization.  Following NVRC, it aims to optimize a neural representation $F$ parametrized by $\theta$ for a given video $V$, using gradient-based optimization - $\theta$ is effectively a representation of $V$. $\theta$ are then compressed by a set of additional parameters $\phi$, which include both entropy model and quantization parameters. Based on NVRC, a set of lightweight parameters $\psi$ is further used for coding $\phi$, forming a hierarchical parameter coding structure. In this case, the encoding process can be formulated as an optimization problem with the following rate-distortion loss:
\begin{equation}
\mathop{\mathrm{argmin}}_{\theta, \phi, \psi}L = \mathop{\mathrm{argmin}}_{\theta, \phi, \psi}(R + \lambda D),
\label{eq:rd}
\end{equation}
where $D$ stands for the reconstruction distortion, e.g., the MSE loss, and $\lambda$ controls the trade-off between rate and distortion, allowing the model to match different rates by simply adjusting $\lambda$. $R$ is the rate loss, which can be estimated by:
\begin{equation}
    R = -\sum^{|\hat{\theta}|}_{n} log_{2}(p_{\hat{\phi}}(\hat{\theta}[n]))
    -\sum^{|\hat{\phi}|}_{n} log_{2}(p_{\hat{\psi}}(\hat{\phi}[n])),
\end{equation}
in which $\hat{\theta}$, $\hat{\phi}$ and $\hat{\psi}$ represent the corresponding quantized parameters. In practice, we also scale $R$ by the video resolution, i.e., by $\frac{1}{T \times H \times W}$, for estimating the bitrate in bits-per-pixel. For simplicity, we omit the description of the highest‐level compression models and their parameter $\psi$ in what follows. These can be found in the original NRVC literature \cite{kwan2024nvrc}.

To reduce the computational complexity of NVRC in both encoding and decoding, in this work we utilize a low complexity neural representation, which can achieve efficient coding ($<$10kMAC/pixel) (Section \ref{subsec:representation}). It should be noted that, while existing low complexity neural representations have already achieved this level of computational complexity, they typically do not meet the requirements of real-world decoding speed. This is due to the use of autoregressive models, which perform coding with a larger number of steps. Thus, we employ an octree-based model, exploiting the spatial-temporal dimension of the feature grid tensors, which enables coding in just a few steps (Section \ref{subsec:entropymodel}).

\subsection{Multi-scale Neural Representation}\label{subsec:representation}

Recent INR-based video coding algorithms can be classified into two categories: (i) those based on COOL-CHIC \cite{leguay2024cool}, which was originally developed for image coding and reconstructs the target signal by performing pixel-wise mapping; (ii) those based on NeRV \cite{chen2021nerv}, typically performing a frame-based or patch-based mapping. The former focuses on extremely low complexity cases, while the latter aims to achieve the best rate-distortion performance. Although both classes utilize learned synthesis transforms \cite{leguay2024cool,chen2021nerv}, which map the latent code to the decoded frame, a key difference is how the computation is implemented - COOL-CHIC–based codecs perform most of the computation, i.e., the synthesis transforms, at the output resolution, even though their latent code is stored in a multi-resolution format. In contrast, NeRV-style representations perform these transforms at multiple resolutions. 

This incurs a trade‑off between parameter count and computational complexity. Consider a multi‑resolution representation with $N$ stages of layers, where the number of channels at each stage is $C,\;\frac{C}{2},\;\frac{C}{4},\;\dots,\;\frac{C}{2^{N-1}}$ and the corresponding spatial resolutions are
$\frac{H}{2^{N-1}}\times\frac{W}{2^{N-1}},\;\dots,\;\frac{H}{2}\times\frac{W}{2},\;H\times W$, respectively. A linear layer\footnote{The same applies to each convolutional layer with a constant kernel size.} at stage $n$ therefore has a cost proportional to $\left(\frac{H}{2^{N - n - 1}}\right)\left(\frac{W}{2^{N - n -1}}\right)\left(\frac{C}{2^{n - 1}}\right)^{2}
\propto H\cdot W\cdot C^{2}$, i.e., the cost remains constant across all stages. By performing part of the computations in a multi‑resolution fashion, features are effectively shared between neighbouring pixels, which typically yields better reconstruction quality for the same computational budget. However, executing an equivalent amount of computation at lower resolutions also requires a larger number of parameters, as the number of parameters is $\left(\frac{C}{2^{n - 1}}\right)^{2}$. When the parameter count is limited, such as in the case of employing video representations for compression, utilizing the parameter budget on high resolution computation may yield better performance, thus limiting the use of multi-resolution features in a low complexity codec. However, by efficiently encoding the parameters to mitigate the associated bit overhead, it is possible to achieve improved performance with multi‑resolution representations \cite{kwan2023hinerv,gao2025pnvc}. This can be achieved through using an end‑to‑end optimization framework, such as NVRC \cite{kwan2024nvrc}.

Selecting the NeRV-style video representation in this work, we propose using HiNeRV \cite{kwan2023hinerv} (as used in the original NVRC \cite{kwan2024nvrc}) but with modifications tailored to the ultra‑lightweight configuration. Specifically, we feed HiNeRV with multiple feature grids at different resolution stages\footnote{Note that the original HiNeRV also use multi-resolution grids, but the grid features where only applied to the first stage of the network.}. This strategy parallels COOL-CHIC-style models and produces significant performance gains when the model’s capacity for feature upscaling (as in the original HiNeRV) is limited, especially for high bitrate regime. We noted that existing NeRV-based models typically use larger models for higher quality rate points. Our modification thus creates a hybrid architecture that combines the strengths of both approaches. To reduce the computational complexity, we simply reduce the number of HiNeRV blocks and the number of channels, and replace the 3D convolutional stem with an addition of a 2D and a 1D convolutional layer in the spatial and temporal dimensions, respectively. With the resulting multi-scale HiNeRV, we achieve efficient video coding compared to existing lightweight models, requiring only 7kMACs/pixel for coding at different rates, where the original HiNeRV consumes at least 170kMACs/pixel.

\subsection{Octree-based Entropy Model}\label{subsec:entropymodel}

\begin{figure}[!t]
    \centering
        \includegraphics[width=0.9\linewidth]{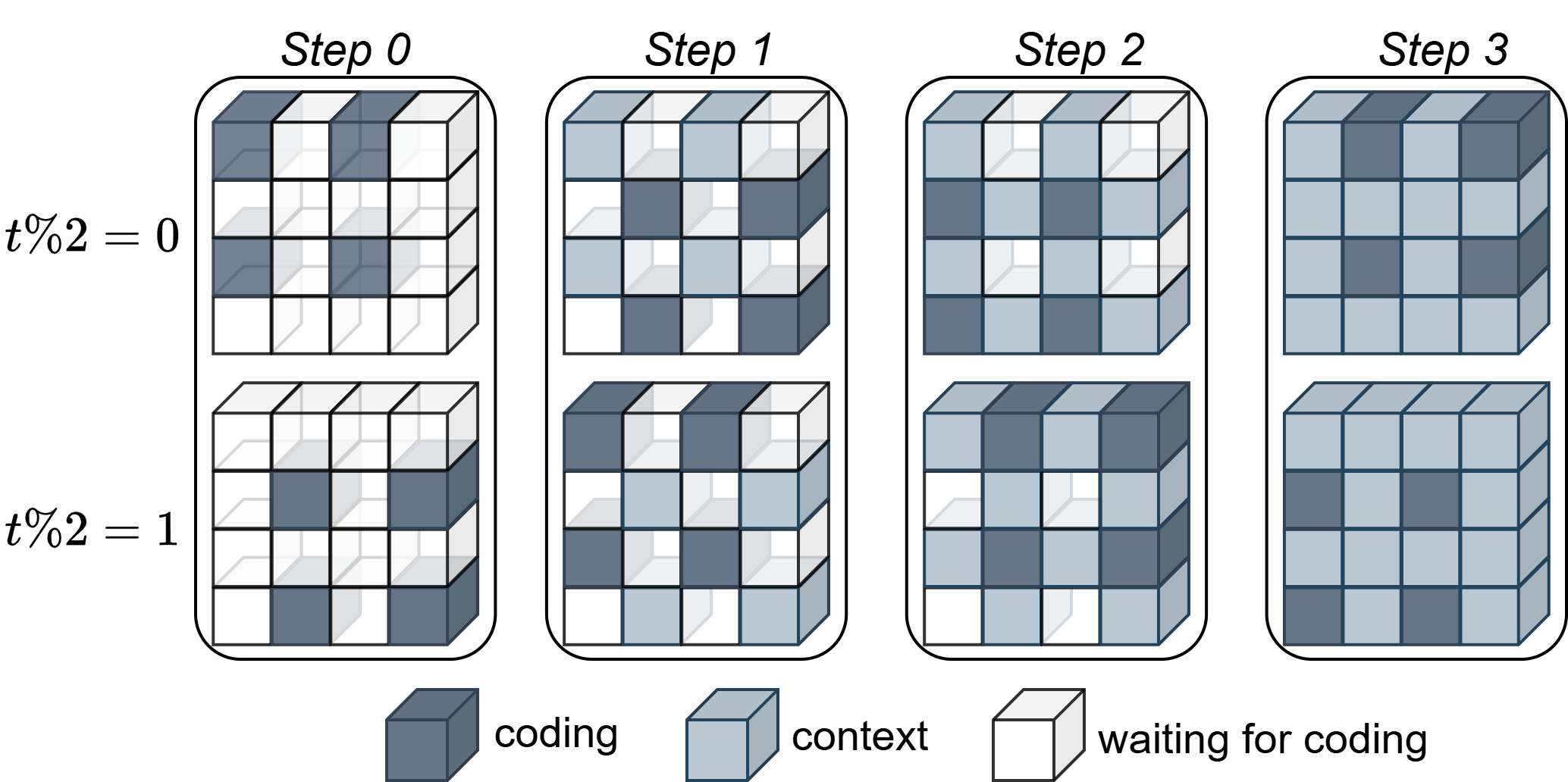}
    \caption{The proposed octree-based entropy coding structure. Here, $t$ represents the feature index in the temporal dimension. We code both the odd and even indexed features in every coding step to reduce coding complexity.}
    \label{fig:octree}
    \vspace{-5pt}
\end{figure}

Existing neural video representations typically use variants of an autoregressive entropy model for coding feature grids, i.e., for coding a 3D latent $z \in \mathbb{R}^{T_{\mathrm{grid}}\times H_{\mathrm{grid}}\times W_{\mathrm{grid}}}$ (omitting the channel dimension here for simplicity), the entropy model estimates $p(z_{t, h, w}|z_{<t,h,w}, z_{t, <h, w}, z_{t, h, <w})$ when coding the value indexed at $(t, h, w)$. This approach, while providing strong coding performance, could limit the practical applications of these low-complexity codecs due to the slow entropy decoding speed resulting from the sequential decoding process, as it yields $T_{\mathrm{grid}}\times H_{\mathrm{grid}}\times W_{\mathrm{grid}}$ coding steps. 
It is noted that some recent works started using more efficient coding models, which partition the space using different patterns such as the checkerboard-based \cite{he2021checkerboard} or quadtree-based \cite{li2023neural} schemes, and perform coding in only a few steps.

Inspired by these efficient models, we designed an octree‑based entropy model for feature‑grid coding. This model can be considered as a 3D extension of checkerboard‑ or quadtree‑based schemes. Specifically, given a feature‑grid tensor 
$z \in \mathbb{R}^{T_{\mathrm{grid}}\times H_{\mathrm{grid}}\times W_{\mathrm{grid}}}$ (channel dimension is omitted),
we partition its spatio‑temporal dimensions based on an octree structure. For each $2\times2\times2$ sub‑block in the tensor, our octree‑based entropy model codes a subset of values in each block in one step. Since the previously coded latents can provide context, which provides a better estimation of the remaining codes, our entropy model models $p(z_{t, h, w}|m_{k} \otimes z)$ as a conditional Gaussian distribution for coding, where $m_{k}$ denotes the mask pattern, which solely depends on the coding step $k$. The detail of the masking scheme is shown in \autoref{fig:octree}.

When applying an entropy model with multiple coding steps directly to the feature grids, the high GPU memory requirement could be an issue, as the entire 4D grid tensor must be held and optimized at once.  Autoregressive models can mitigate this via masked convolutions, but such a masking process is incompatible with our octree‑based scheme. Consequently, spanning eight sequential coding steps becomes prohibitively costly in terms of both memory and runtime. To address this issue, we propose to code two values in every $2\times2\times2$ sub‑block in the tensor in each step, reducing the total number of coding steps to four. With the high dimensional feature grids we used (up to $96\times540\times960\times1$ in this work), our entropy model can still fit into GPU for efficient coding.

In addition to the octree‑based spatio‑temporal partitioning, our model interleaves the coding order across channels — drawing inspiration from the existing spatial prior schemes \cite{li2022hybrid, li2023neural}. We further optimize a block‑wise quantization step size and introduce a block‑wise auxiliary conditional latent, both following the block‑partitioning scheme of NVRC~\cite{kwan2024nvrc}. The conditional latent is learned end‑to‑end and fed into the entropy model to better condition the local distribution of the latent code, analogous to the Hyperprior \cite{ballé2018variational} approach commonly used in neural image and video compression.  

\section{Results and Discussion}\label{sec:experiment}

We evaluated the proposed NVRC-Lite on HD (1080p) test sequences from the UVG \cite{mercat2020uvg} and HEVC-B \cite{bossen2010common} datasets, using 96 frames per sequence converted from the original YUV 4:2:0 colorspace to RGB 4:4:4, based on the BT.601 standard using FFmpeg, following \cite{kwan2024nvrc}. Benchmarks include one conventional codec x265 \cite{x265} with both the \textit{medium} and \textit{veryslow} presets and one lightweight INR-based video codec C3 \cite{kim2024c3}. For C3, our experiment follows the default configuration provided in its official open-source implementation \cite{kim2024c3}, except that we employed a shorter training time (30k steps) which can already yield a comparable performance to the full training (as suggested in \cite{kim2024c3}). The adaptive setting was omitted as it was unavailable in the provided code and is prohibitively time-consuming, requiring a sweep of nine hyperparameters per patch. We have excluded the closely related baselines HiNeRV \cite{kwan2023hinerv} and NVRC \cite{kwan2024nvrc} to maintain a clear focus on low-complexity video compression. We measure the fidelity of reconstructed videos using both PSNR and MS-SSIM \cite{wang2003multiscale}. To compare overall compression efficiency, we report the Bj{\o}ntegaard Delta Rate (BD-rate) \cite{bdrate} of the proposed NVRC-Lite against each benchmark codec.

To train NVRC-Lite, we adapted the optimization pipeline from NVRC \cite{kwan2024nvrc} by using 1440/120 epochs for the two stages to suit shorter sequences, setting the batch size to 4 for improved GPU utilization, and using the MSE loss as the distortion term.

\subsection{Compression Performance}

\begin{figure}[!t]
  \centering
  \vspace{-3pt}
  \begin{subfigure}[b]{0.485\linewidth}
    \centering
    \includegraphics[width=\linewidth]{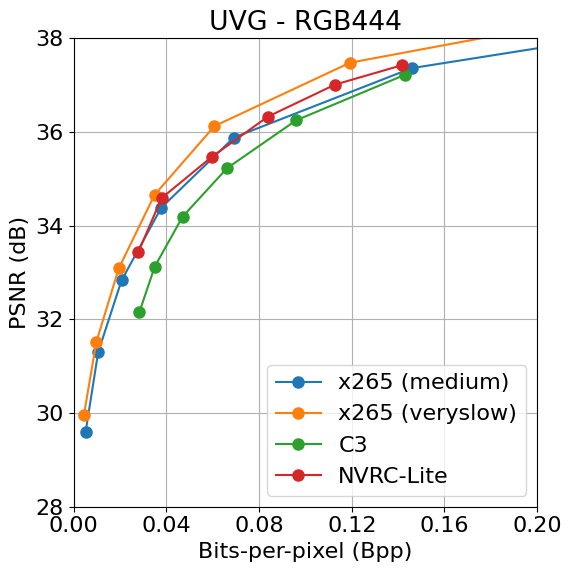}
    \label{fig:rd-plot-uvg}    
  \end{subfigure}
  \begin{subfigure}[b]{0.485\linewidth}
    \centering
    \includegraphics[width=\linewidth]{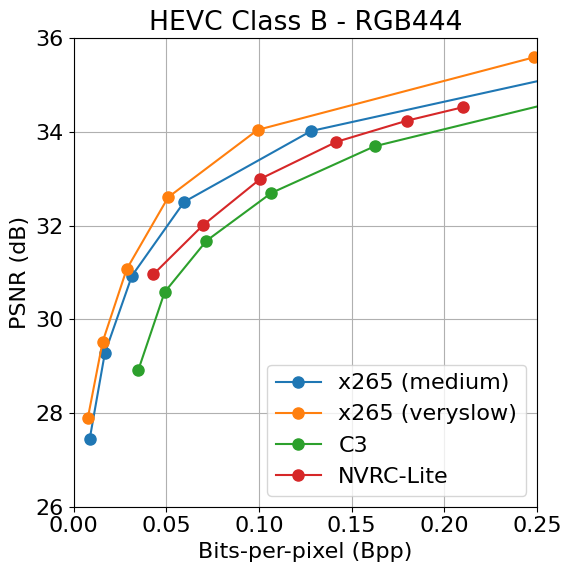}
    \label{fig:rd-plot-hevc}   
  \end{subfigure}
  \vspace{-10pt}
  \caption{Rate–distortion curves on the UVG and HEVC‑B datasets.}
  \label{fig:rd-plot}
\end{figure}

\begin{figure}[!t]
\small
    \centering
    \begin{minipage}{0.24\textwidth}
        \centering
        \includegraphics[width=\linewidth]{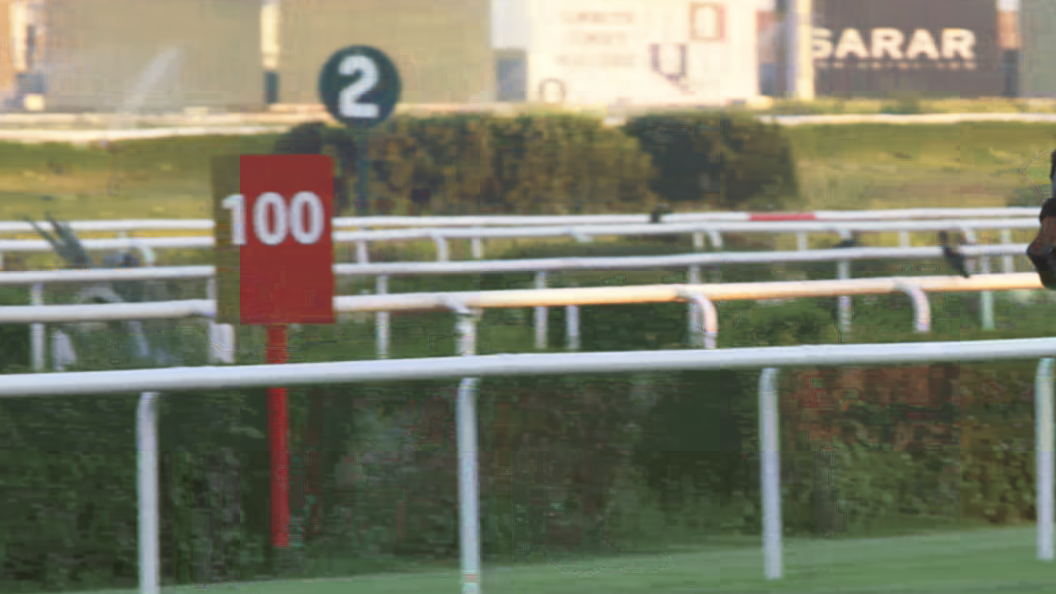}
        34.23 dB PSNR, 0.1088 bpp 
    \end{minipage}
    \hfill
    \begin{minipage}{0.24\textwidth}
        \centering
        \includegraphics[width=\linewidth]{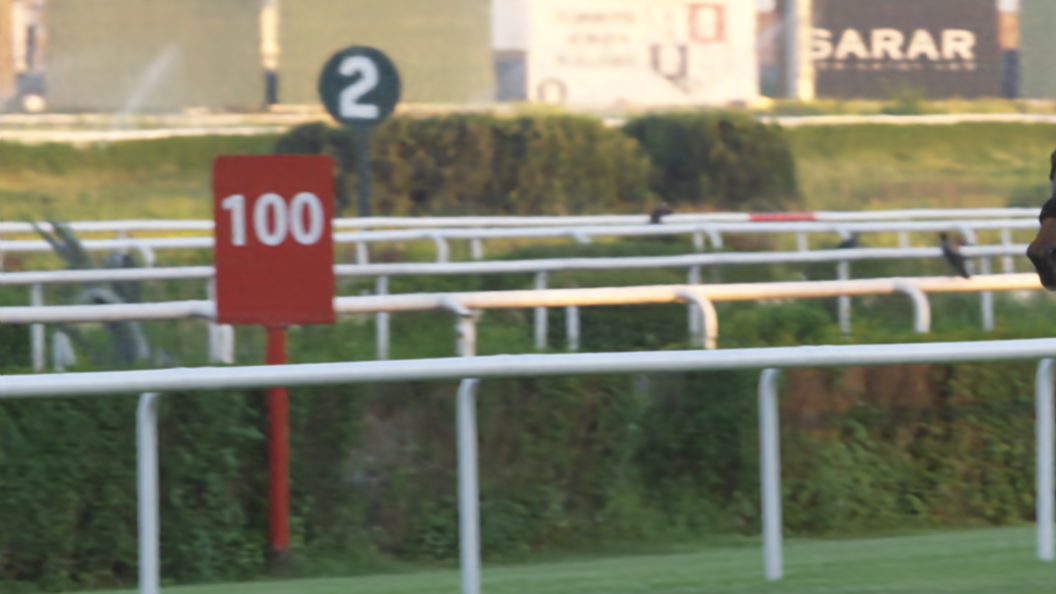}
        36.43 dB PSNR, 0.0884 bpp 
    \end{minipage}
    \begin{minipage}{0.24\textwidth}
        \centering
        \includegraphics[width=\linewidth]{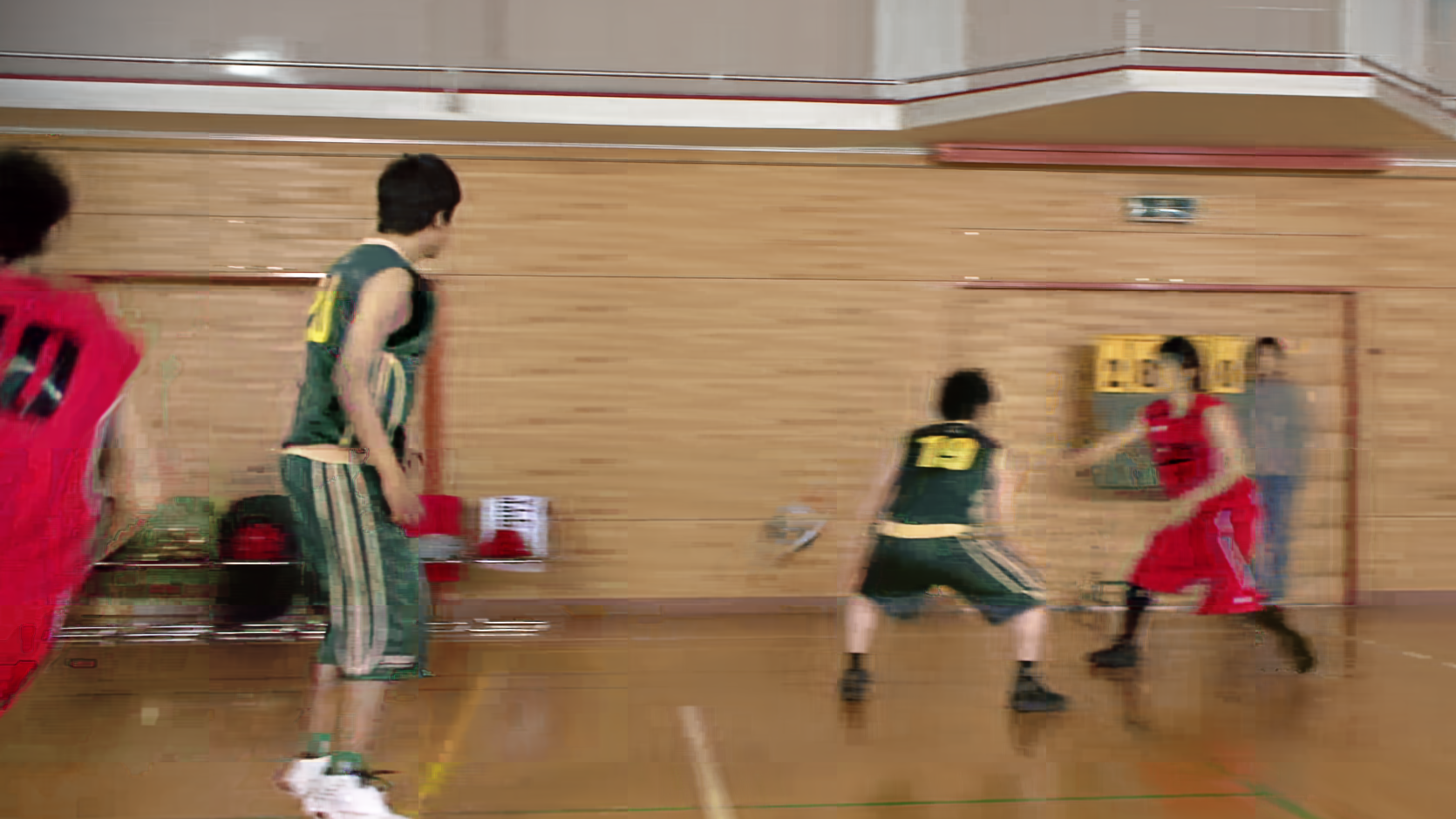}
        31.62 dB PSNR, 0.0720 bpp 
    \end{minipage}
    \hfill
    \begin{minipage}{0.24\textwidth}
        \centering
        \includegraphics[width=\linewidth]{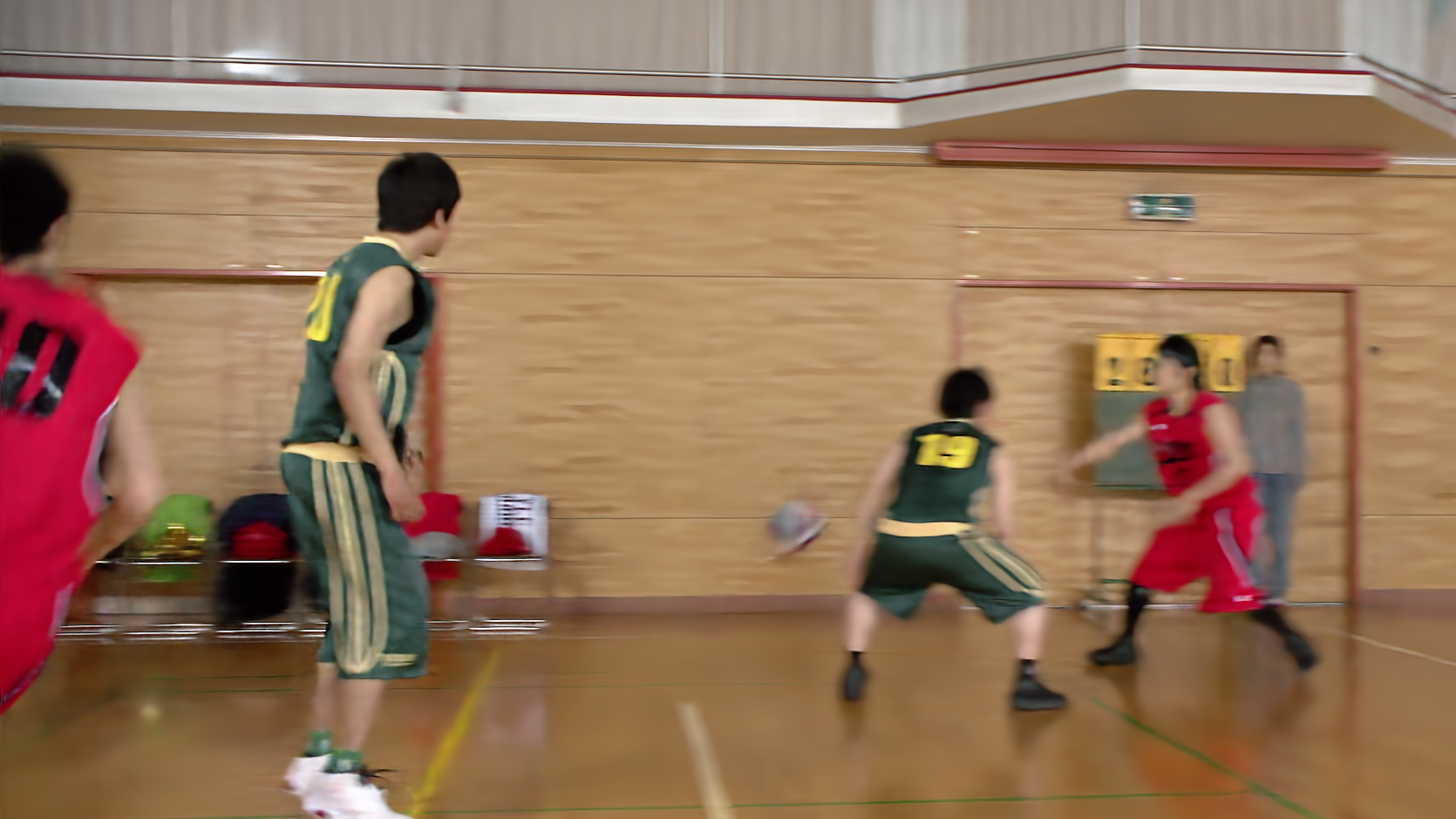}
        33.12 dB PSNR, 0.0552 bpp 
    \end{minipage}
    \hfill
    \caption{Examples of visual comparison between C3 (left) and our method (right). Examples are from the UVG (top) and HEVC-B (bottom) datasets.}
    \label{fig:visual}
    \vspace{-10pt}
\end{figure}

The rate-distortion (RD) performance of the proposed NVRC-Lite codec is shown in \autoref{tab:BD-rate} and \autoref{fig:rd-plot}. Compared to the SOTA low-complexity codec C3, NVRC-Lite offers average bitrate savings of 18.54\% in PSNR and 21.23\% in MS-SSIM, and delivers up to 5.27\% coding gain in PSNR over x265-\textit{medium}. We also provide visual comparisons between example frames reconstructed by C3 and the proposed NVRC-Lite in \autoref{fig:visual}. It can be observed that our model produces visually superior results with finer textural details and fewer visible artifacts under a tighter bitrate budget. 

\begin{table}[!t]
    \centering
    \caption{BD-rate results with the UVG and HEVC (Class B) datasets. The baseline codec in each column is used as the anchor.} \vspace{-4pt}
    \label{tab:BD-rate}
    \resizebox{1\linewidth}{!}{%
    \begin{tabular}{@{} rr cccc @{}}
        \toprule
        Dataset & Metric & x265 (\textit{medium}) & x265 (\textit{veryslow}) & C3 \cite{kim2024c3}\\
        \midrule
        \multirow{2}{*}{UVG} & PSNR & -5.27\% & 17.02\% & -21.03\% \\
        & MS-SSIM & 3.35\% & 20.18\% & -23.06\% \\
        \cdashline{1-6} \addlinespace
        \multirow{2}{*}{HEVC-B} & PSNR & 28.34\% & 56.74\% & -16.05\% \\
        & MS-SSIM & 16.36\% & 42.00\% & -19.40\% \\
        \bottomrule
    \end{tabular}%
    }
\end{table}

\begin{table}[!t]
    \centering
    \caption{Computational complexity comparison. 
    The quoted numbers are the speed with entropy coding. It is noted that our current entropy coder is unoptimized, and C3's implementation does not perform actual entropy coding.} \vspace{-4pt}
    \label{tab:complexity}
    \resizebox{0.85\linewidth}{!}{%
    \begin{tabular}{rcccc}
        \toprule
        Model & 
        & kMACs/pixel & Enc. FPS & Dec. FPS \\
        \midrule
        C3 \cite{kim2024c3}& 
        & 4.4 & 0.0007 & 38.09 (n/a) \\
        NVRC-Lite & 
        & 7.8 & 0.0059 & 97.28 (4.86) \\
        \bottomrule
    \end{tabular}%
    }
\end{table}

\begin{table}[!t]
    \centering
    \caption{Ablation study. Results are conducted in three sequences (Beauty/Jockey/ReadySetGo) from the UVG dataset, and BD-rate with PSNR/MS-SSIM are reported.} \vspace{-4pt}
    \resizebox{1\linewidth}{!}{%
    \begin{tabular}{cccc}
        \toprule
         Variants     & Baseline  & w/ Autoregressive & w/ Single scale \\
         \midrule
         BD-rate (\%) & 0.0\%/0.0\% & 9.78\%/13.92\% & -6.27\%/-17.68\% \\
         \bottomrule
    \end{tabular}
    }
    \label{tab:ablation}
    \vspace{-10pt}
\end{table}

\subsection{Complexity Comparison}
As computational complexity is a primary focus of this work, we also profile C3 and NVRC-Lite based on their MACs/pixel, encoding time, and decoding time. All these complexity figures were estimated on a single NVIDIA RTX4090 GPU for 1080p frames. The model compression MACs and encoding/decoding time are measured in terms of the steps for performing quantization and entropy coding. As shown in \autoref{tab:complexity}, despite larger MACs/pixel values, the proposed NVRC-Lite is substantially faster, achieving an 8.4$\times$ speedup in encoding and a 2.5$\times$ speedup in decoding compared to C3.

\subsection{Ablation Study}

We provide ablation study results in \autoref{tab:ablation}, for the variant which replaces the proposed octree-based entropy model with the autoregressive model \cite{kwan2024nvrc}, and that using the single resolution grid feature design as in NVRC \cite{kwan2024nvrc}. Our study found that, our octree-based entropy model has outperformed autoregressive-based model, and it is based on unoptimized code, but has already sped up the entropy coding time by $\sim$4 times compared to the autoregressive model. We also found that with single scale grid features, the BD-rate is increased, however, the model cover a much narrower quality range (around half of the baseline).

\section{Conclusion}\label{sec:conclusion}
In this paper, we propose NVRC-Lite, a lightweight neural video representation compression framework that leverages multi-scale and higher resolution feature grids to enhance the performance of implicit neural representations (INRs) at low complexity. An octree-based entropy coding module has been proposed to further accelerate compression and improve system efficiency. The results show that our method achieves notable BD-rate savings and reduced encoding and decoding complexities compared to the state-of-the-art lightweight INR-based codec, C3, with even lower encoding and decoding computational complexities. In future work, we will focus on further improving the compression performance of low-complexity INR codecs and extending their applicability to more diverse settings such as the low delay coding case. 


\small
\bibliographystyle{IEEEtran}
\bibliography{Ref}

\end{document}